\documentclass[runningheads]{llncs}

\usepackage{graphicx}
\usepackage{booktabs}
\usepackage{times}
\usepackage{epsfig}
\usepackage{graphicx}
\usepackage{amsmath}
\usepackage{amssymb}
\usepackage{booktabs}
\usepackage{subcaption}
\usepackage{tabularx}
\usepackage{wrapfig}
\usepackage{footnote}
\makesavenoteenv{tabular}
\makesavenoteenv{table}

\usepackage[hyphens]{url}

\usepackage[accsupp]{axessibility}
\newcommand{\name}{S4} 
\newcommand{\datasetname}{m2s2-SITS} 

\newcommand{\para}[1]{\vspace{3pt}\noindent\textbf{#1}}
\newcommand{\squishlist}
{
    \begin{list}{$\bullet$}
    {
        \setlength{\itemsep}{0pt}      \setlength{\parsep}{2pt}
        \setlength{\topsep}{2pt}       \setlength{\partopsep}{0pt}
        \setlength{\leftmargin}{1em} \setlength{\labelwidth}{0.5em}
        \setlength{\labelsep}{0.5em}
    }
}
\newcommand{\squishend}
{
    \end{list}
}

\usepackage[pagebackref,breaklinks,colorlinks]{hyperref}

\begin{document}

\title{S4: Self-Supervised Sensing Across the Spectrum}

\titlerunning{S4}

\author{Jayanth Shenoy, Xingjian Davis Zhang, Bill Tao, Shlok Mehrotra, Rem Yang, Han Zhao, Deepak Vasisht}

\authorrunning{Shenoy et. al}

\institute{University of Illinois Urbana-Champaign}

\maketitle

\begin{abstract}
Satellite image time series (SITS) segmentation is crucial for many applications like environmental monitoring, land cover mapping and agricultural crop type classification. However, training models for SITS segmentation remains a challenging task due to the lack of abundant training data, which requires fine grained annotation. We propose \name\ a new self-supervised pre-training approach that significantly reduces the requirement for labeled training data by utilizing two new insights: (a) Satellites capture images in different parts of the spectrum such as radio frequencies, and visible frequencies. (b) Satellite imagery is geo-registered allowing for fine-grained spatial alignment. We use these insights to formulate pre-training tasks in \name. We also curate \datasetname, a large-scale dataset of unlabeled, spatially-aligned, multi-modal and geographic specific SITS that serves as representative pre-training data for \name. Finally, we evaluate \name\ on multiple SITS segmentation datasets and demonstrate its efficacy against competing baselines while using limited labeled data. 
\end{abstract}

%
%
\newcommand{\red}[1]{{\color{red}#1}}
\newcommand{\todo}[1]{{\color{red}#1}}
\newcommand{\TODO}[1]{\textbf{\color{red}[TODO: #1]}}
\newcommand{\bt}[1]{{\color{blue}[BT: #1]}}

\section{Introduction}
\begin{figure*}[ht]
    \centering
    \includegraphics[width=\linewidth]{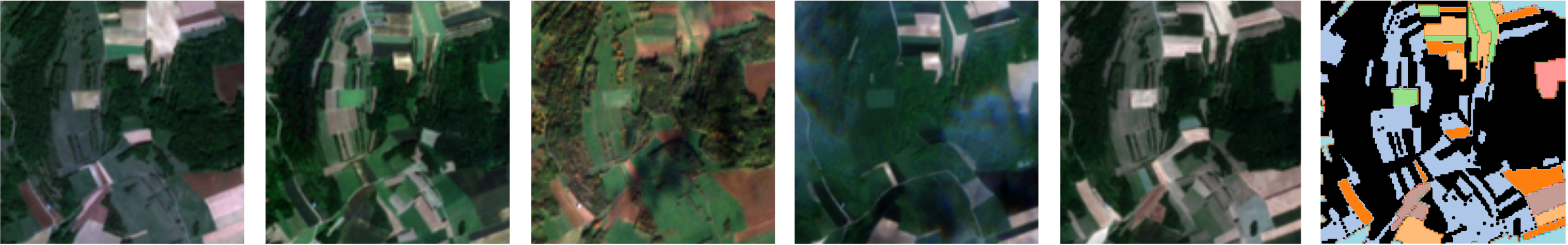}\vspace{-0.1in}
    \hspace{.03\textwidth}\vspace{-0.1in}
  \caption{Optical images in one SITS captured at different points in time over the same location. The rightmost image is the segmentation mask corresponding to this spatial location. The different images illustrate the significant temporal variation that occurs during crop growth. }
\label{tsexample}\vspace{-0.1in}
\end{figure*}

In recent years, many organizations~\cite{planetlabs,spireglobal,capella} have launched large satellite constellations for Earth observation. These constellations regularly capture high-resolution Earth imagery that is critical for measuring climate change \cite{climate_change,plasticai}, responding to humanitarian crises \cite{humancris}, precision agriculture \cite{agriculture}, and natural resources management \cite{naturalresource}. Specifically, satellites with multiple visits over a given location on Earth provide unique insights into complex spatial and temporal patterns~\cite{panop_segmentation,sits1,sits2} at such locations, unlike single satellite images. These Satellite Image Time Series (SITS) (as shown in Fig.~\ref{tsexample}), for example, can provide greater insights into how crops on a farm grow over time, what types of crops are growing, or when the crops are ready to be harvested. SITS is also more robust to temporary disruptions such as cloud cover that may occur in single images. Due to its key advantages and environment implications, SITS semantic segmentation has become a task of critical importance and has widespread use in many Earth sensing applications such as deforestation monitoring \cite{deforestation}, urban planning \cite{urbanplanning}, and agriculture crop type classification \cite{africacrop}.

However, training segmentation models for SITS requires collecting large amounts of labeled data, requiring laborious manual annotation from domain experts \cite{raechelabianchetti}. This is especially challenging for semantic segmentation which requires pixel-level annotations. Moreover, many satellite images use non-optical channels \cite{multispec} beyond the standard RGB wavelength, making them difficult to interpret for humans. 

We propose \name, a novel \textit{self-supervised approach} for semantic segmentation of SITS that eliminates the need for large amounts of labeled data. We observe that while labeling requires human effort, unlabeled data is abundant because satellites continuously orbit the Earth and collect data. Our key insight is that we can leverage this unlabeled data by utilizing two properties unique to SITS:
\squishlist
\item \textbf{Multi-modal Imagery:} Different satellites (or different sensors on the same satellite) collect images in different parts of the electromagnetic spectrum (e.g. RGB, radar). We can use such multi-modal images for cross-modal self-supervision.
\item \textbf{Spatial Alignment and Geographic Location:} Satellite images are geo-referenced, i.e., each pixel has a geographic coordinate (latitude and longitude) associated with it. This allows for spatial alignment between data collected in different parts of the spectrum. 
\squishend

Although prior self-supervised vision solutions exist, they largely do not leverage the geographical, temporal, and multi-channel characteristics of SITS. Prior self-supervised solutions are mainly designed for natural images that consist of only optical RGB channels \cite{simclr,cl2,cl3}. While more specialized self-supervised methods for satellite imagery exist, they are largely either single-modal \cite{satmae,caco}, or mono-temporal \cite{seco,gssl}, and often only consider a subset of the characteristics of SITS mentioned above. As a result, prior work does not appropriately learn the underlying representations required for SITS segmentation.

Given the unique properties of SITS, \name\ exploits the abundant unlabeled satellite data through cross-modal self-supervision. Specifically, we use different data modalities for a given location to learn informative intermediate representations \emph{without any labeled data}. Using unlabeled SITS, we can pre-train representative SITS encoders that perform effectively on downstream SITS segmentation. We achieve this by pre-training SITS segmentation models through two auxiliary tasks:
\squishlist
\item \textbf{Cross-Modal Reconstruction Network:}~We design a new cross-modal SITS reconstruction network that attempts to reconstruct imagery in one modality (e.g. radar) from the corresponding imagery in another modality (e.g. optical). Our reconstruction network encourages the encoder networks to learn meaningful intermediate representations for pixel-wise tasks by leveraging the structured spatial alignment in satellite image data.

\item \textbf{MMST Contrastive Learning:}~We formulate a novel multi-modal, spatio-temporal (MMST) contrastive learning framework for SITS. We train one encoder for each modality (e.g. for radar and optical imagery) and align the intermediate representations using a contrastive loss. Our contrastive loss operates along both the space and time dimensions of the feature space to align multi-modal SITS. Intuitively, our loss helps negate the impact of temporary noise (such as cloud cover) that is visible in only one of the input images. 
\squishend

Our design solves multiple challenges unique to SITS segmentation. First, \name\ can significantly reduce the need for labeled training data by exploiting abundant unlabeled images. Second, some image modalities such as optical images are obstructed by clouds, leading to missing or incorrect information (see Fig.~\ref{cloudexample}). In fact, around 75\% of the Earth's surface is covered by clouds at any given point in time \cite{TwoYearsofCloudCoverStatisticsUsingVAS,AssessmentofGlobalCloudDatasetsfromSatellitesProjectandDatabaseInitiatedbytheGEWEXRadiationPanel,zhao2021seeing}. Through pixel-wise alignment of radar and optical image encoders, \name\ effectively leverages the rich information provided by radar in cloudy settings (since radar passes through clouds). As a result, \name\ pre-trains powerful encoders for both modalities that can reduce the negative impact of cloud cover on model performance. Third, for each modality, the reflectance value of each pixel is different. Because of this, certain Earth surface characteristics may be clearly visible in one modality, but not necessarily as visible in the other. \name\ solves this through its cross-modal reconstruction network which is able to infer the presence of vegetation (for example) in one modality based on the patterns learned from the other. Intuitively, this enables the model to learn to understand the signatures or indicators that would correspond to certain features across modalities.

Importantly, S4 delivers \textit{single-modality inference}. Single-modality inference is crucial due to two real-world constraints. First, satellites capturing images of different modalities may be operated by different entities. In fact, 95\% of Earth Observation satellites are equipped with only a single sensing modality~\cite{UCSSatelliteDatabase}. Thus, while multi-modal training data may be available through aggregating public datasets, such data is almost always not available at inference time \cite{radaropt}. Second, requiring both modalities during inference increases the delay of decision-making in response to critical events (e.g. floods and fires), since multiple modalities can be offset in time by several hours to days (depending on satellite orbits) \cite{l2d2,serval,umbra}. Hence, while we leverage multi-modal data at training time, we limit ourselves to a single modality inference.

\begin{figure*}[t]
    \centering
    \includegraphics[width=.9\linewidth]{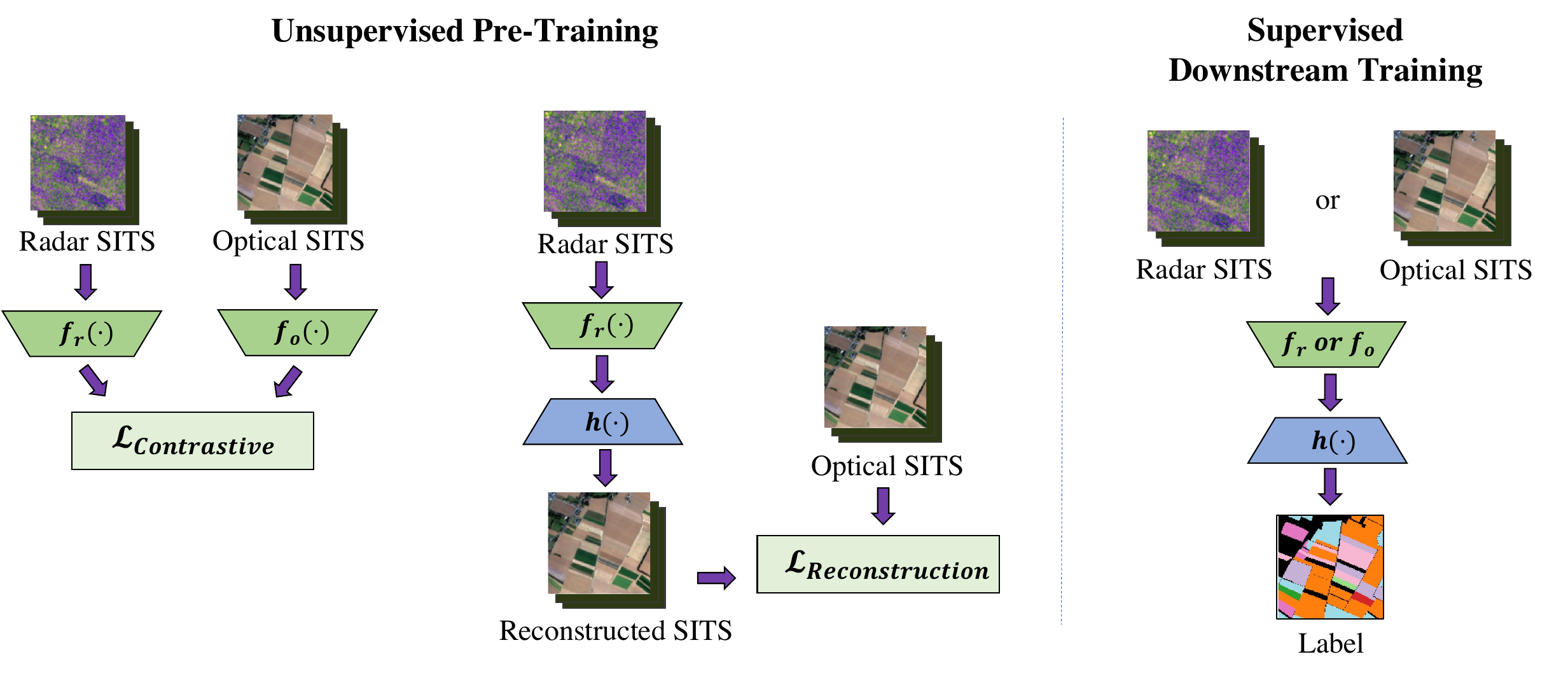}\vspace{-0.1in}
    \caption{\textbf{Overview of \name.} \name\ takes in temporally pre-processed multi-modal time series data. During pre-training, radar-optical SITS pairs flow through the network and our proposed MMST contrastive loss and Cross-Modal reconstructive loss operate on their encodings. After pre-training, a small amount of labeled data is used to fine-tune the model for SITS segmentation.}
    \label{fig:overview}
    \vspace{-0.2in}
\end{figure*}

We evaluate \name's performance on two satellite image datasets: PASTIS and Africa Crop Type Mapping segmentation tasks. To demonstrate the efficacy of aligned satellite imagery and showcase the opportunity for self-supervised pre-training for SITS segmentation, we collected \datasetname, two large-scale unlabeled but modality-aligned datasets of satellite images corresponding to the same regions of our labeled datasets. In our evaluation, we pre-train our model using our curated dataset \datasetname\ and finetune the models on the existing datasets with segmentation labels. Experiments demonstrate that \name\ outperforms competing  self-supervised baselines for segmentation, especially in the case where the number of labeled data is relatively small. As a result, \name\ takes a first step towards self-supervised SITS segmentation through novel techniques that reliably leverage multi-modal and spatially aligned imagery. In summary, this paper makes the following contributions:
\squishlist
\item We propose \name, a self-supervised training method for SITS segmentation that considers the unique structural characteristics of satellite data such as multiple modalities, spatial alignment, and temporal change through novel cross-modal reconstruction and contrastive learning frameworks

\item We release \datasetname, a large dataset of spatially-aligned, multi-modal SITS to aid in self-supervised pre-training.

\item We evaluate \name\ on multiple SITS datasets and benchmark our approach against other self-supervised approaches commonly used on satellite imagery. Our results demonstrate the effectiveness of \name\ through significant improvement over prior state-of-the-art methods on downstream SITS segmentation.
\squishend
\vspace{-0.2in}
\section{Related Work}
\vspace{-0.2in}
To the best of our knowledge, we present the first self-supervised approach for semantic segmentation in multi-modal SITS. We discuss related work below.

\para{Learning with Satellite Imagery} Prior work on satellite imagery can be characterized as (a) single-image or (b) SITS. Single-image methods \cite{transcoding,lockheed_sat_self_supervised,Zhang_2022_CVPR}, although more extensively studied, are unable to effectively gain insights into many environmental sensing applications that typically evolve over time, such as crop mapping and disaster monitoring \cite{ag_cvpr,rudner2018multimathbf3net}. Most of the prior work in self-supervised learning for satellite imagery are single-image unimodal techniques \cite{seco,gssl} that cannot leverage the multi-temporal and multi-modal structure of SITS data. Therefore, there has been a growing number of recent supervised efforts that leverage SITS, which better captures the complex characteristics that evolve over time in many environmental sensing tasks. These efforts have designed SITS-based models for a variety of downstream Earth observation tasks, such as image classification \cite{ag_cvpr}, super-resolution \cite{he2021spatial}, and segmentation \cite{panop_segmentation}. Each of the cited methods focuses on developing space-time encoding for effective feature extraction.

Although the state-of-the-art SITS-based techniques yield vast improvements over single-image methods for a variety of tasks, they mainly rely on unimodal, optical satellite imaging.  However, optical imagery is not robust under low visibility conditions (e.g., due to rain, night, or clouds), making it difficult to obtain such data in time-sensitive settings \cite{ramsauer_2020}. \name\ extracts insights even from non-optical SITS during training, making it significantly more practical in these challenging conditions. 

\para{Learning with Multiple Modalities}
Many modern satellites are equipped with non-optical sensing modalities \cite{sentinal,capella}. Computer vision on non-optical imaging modalities, such as radar, has been explored much less than optical imaging modalities. This is due to radar images being difficult to interpret by humans compared to optical images, making it harder to acquire labeled data. Most prior works focus on exploring radar images using unsupervised techniques \cite{sar1,sar2,sar3}. These techniques do not generalize well to different events and often exhibit limited performance. Prior work on multi-modal satellite imagery has also explored the reconstruction of obscured or cloudy optical images by leveraging aligned non-optical radar images \cite{zhao2021seeing,cloudcover2,cloudcover3}. These multi-modal reconstruction models tend to provide a more accurate optical reconstruction than prior unimodal methods like image in-painting \cite{inpainting1,inpainting2}, demonstrating the potential utility of non-optical multi-modal learning. More recently, there have been efforts to try and incorporate multiple modalities for SITS \cite{rudner2018multimathbf3net,pastis-r,dynamicearthnet}. Such efforts typically focus on designing fusion techniques for modalities along with reliable spatio-temporal encodings to improve performance. \name\ distinguishes itself from these methods by providing a training method that requires significantly less labeled data and only a single SITS modality at inference time.

\para{Self-Supervised Learning}
Self-supervised learning for visual representations has gained prominence within the last few years \cite{ss1,ss2,ss3,ss4,ss5}. One of the most recent notable self-supervised methods has been contrastive representation learning, which attempts to align similar pairs of images as a pre-training task to help with downstream model performance \cite{simclr,cl2,cl3}. Although prior work mainly focuses on instance-level contrastive learning for downstream classification, recent works explore pixel-level contrastive learning techniques, which provide better transfer to segmentation tasks \cite{propself,p2seg,reco,pixcl}. For the majority of these contrastive learning approaches, positive pixel pairs are assigned either by using corresponding pixels with the same label or through corresponding pixels from different augmented views of the same image. In contrast, \name\ leverages the spatial alignment between different satellite modalities and associates positive pairs through corresponding pixels in different modalities. 

\para{Semantic Segmentation of SITS} 
Many prior methods have found success in using UNet-based architectures \cite{Unet} for encoding representations helpful for satellite image segmentation \cite{panop_segmentation,ghanadataset,dynamicearthnet}. More recent efforts specific to SITS have also designed multi-temporal and multi-modal fusion schemes using convolutional encoders \cite{rudner2018multimathbf3net,pastis-r}. The advantage of \name\ is that we require only a single modality of SITS during inference time, whereas every prior multi-modal method requires both. \name\ also incorporates a novel self-supervised approach that significantly reduces the need for labeled data.

\vspace{-0.1in}\section{Problem Setup: Satellite Imaging with \\ Multiple Modalities} \label{background}

\begin{figure}[t]
\centering
  \includegraphics[width=.15\columnwidth]
    {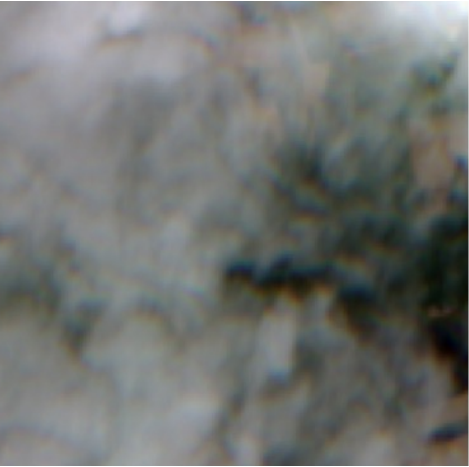}
    \hspace{.06\textwidth}
  \includegraphics[width=.15\columnwidth]
    {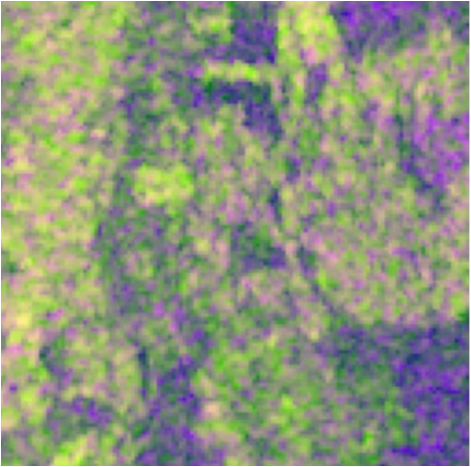}
  \caption{Multi-modal images captured on the same day: while the optical image (left) is occluded by clouds, the radar image (right) is not affected.}
\label{cloudexample}
\vspace{-0.2in}
\end{figure}

Our work is situated in the emerging context where different satellite constellations capture Earth imagery in different frequency bands. We seek to extract spatio-temporal insights from this data. A majority of the satellites capture optical images that passively monitor the reflections of sunlight off the Earth's surface. These optical images are often multi-spectral, including imaging bands outside the standard visible red, green, and blue channels. However, a key disadvantage of such imagery is that optical satellite images are often occluded by clouds (Fig.~\ref{cloudexample}) and are easily obscured in low-lighting conditions, such as night and fog.

Some satellites are equipped with radar imaging that works by actively transmitting pulses of radio waves and measuring the reflectance of these radio pulses to produce radar images. These radio waves utilize a longer wavelength than optical images and are typically better at monitoring certain aspects of the surface, such as moisture and topology. However, the resolution of radar imagery is lower than that of optical images. 
Satellites are typically equipped with either optical or radar imaging modalities, \emph{but not both} \cite{radaropt}. Therefore, images in optical and radar SITS cannot be perfectly aligned in time. 

Each image captured by satellites is georeferenced, i.e., we can extract per pixel geographic coordinates. This allows us to spatially align images even when captured on different satellites. However, leveraging the temporal aspect of SITS data poses some challenges. First, images in SITS, unlike videos, are not taken at regular intervals. Images are taken over a location only when a satellite orbits over that location, meaning the time between images in SITS is irregular based on the satellite's orbit. Second, for multi-modal SITS, different sensing modalities are often located on different satellites, meaning that images of SITS of different modalities are not only unaligned in time, but they can also result in time series of vastly different lengths. 

\name's primary task is to use the ample amount of unlabeled imagery collected by satellites for cross-modal self-supervision. Our formulation builds upon the key idea of \emph{pixel-level semantic consistency between multi-modal images captured over the same location at roughly the same time}. We propose a new training objective that encourages similarity of corresponding space-time features across modalities, while maximizing the distance between the features corresponding to either different locations or different times. Though different modalities have certain distinctions like differing spectral ranges, the \emph{semantic representation of the underlying scene should be agnostic to both wavelength and noise}, e.g., cloud cover (for optical) or capture angle (for radar), and thus \name\ can be used to achieve a course-grained alignment of the multi-modal, multi-temporal features beneficial for self-supervised learning. By leveraging these natural structural characteristics of SITS, our approach extracts a more informative representation that limits the impact of modality-specific noise. 

\para{Notation and Setup}
We consider the respective radar and optical image modalities, $\mathcal{X}_{r} \subset \mathbb{R}^{(T_1 \times C_1 \times H \times W)}$ and $\mathcal{X}_{o} \subset \mathbb{R}^{(T_2 \times C_2 \times H \times W)}$, where $T_i$, $C_i$, $H$, and $W$ are the number of images in the time series, number of image channels, image height, and image width dimensions, respectively. During training, we assume access to $N$ SITS pairs $\{(x^{l_n}_{o}, x^{l_n}_{r})\}^N_{n=1} \in (\mathcal{X}_{o} \times \mathcal{X}_{r})^N$, where $l_i$ corresponds to the location where the SITS was captured. Although we have $N$ total SITS pairs, we assume that only $K$ of these $N$ image pairs ($K \ll N$) have segmentation labels: $\{(x^{l_k}_{o}, x^{l_k}_{r},y^{l_k})\}^K_{k=1}$, where the label $y^{l_k} \in \mathbb{N}^{H \times W}$ maps each pixel location to a given class.

\vspace{-.1in}
\section{Method}\label{sec:method}
\vspace{-.1in}
\para{Overview}

Fig.~\ref{fig:overview} provides an overview of \name. At a high level, \name\ operates in three stages:

\squishlist
    \item \textbf{Pre-Training:} During the pre-training stage, \name\ leverages abundant unlabeled data by jointly optimizing the proposed pixel-wise multi-modal contrastive loss $\mathcal{L}_{c}$ (Sec.~\ref{contrastivesection}) and reconstruction loss $\mathcal{L}_{r}$ (Sec. \ref{reconstructionsection}):
\begin{equation}
\mathcal{L} = \mathcal{L}_{c} + \lambda \cdot \mathcal{L}_{r},
\label{joint_loss}
\end{equation}
where $\lambda$ is a hyperparameter controlling the relative weight between the two loss terms. Neither of the above two losses require labels.
    \item \textbf{Downstream Training:} In this stage, our network is fine-tuned on the $K$ SITS pairs with labels for downstream segmentation by further appending a segmentation module over the learned features (Sec.~\ref{downstreamtraining}). 
    \item \textbf{Inference:} In the final stage, \name\ predicts a single segmentation map per different location from the SITS of a \emph{single} modality (either radar or optical). 
\squishend

\para{Time Series Interpolation} A key challenge for \name\ is that satellites visit the same location at different times, leading to temporal mismatch across modalities. Higher temporal mismatch across images causes more semantic mismatch in the underlying representation. To avoid this problem, we introduce a pre-processing strategy to coarsely align the temporal dimension between differing modalities. This pre-processing step is necessary to ensure finer-grained spatial and time alignment through the rest of the training process.  Recall that we are given as input $x^{l_n}_{r} \in \mathbb{R}^{(T_1 \times C_1 \times H \times W)}$ and  $x^{l_n}_{o} \in \mathbb{R}^{(T_2 \times C_2 \times H \times W)}$, where $T_1 \neq T_2$ in general. We determine which SITS modality has fewer time frames: let $T_{min} = \min(T_1, T_2)$ and define $x^{l_n}_{min} := x^{l_n}_{r}$ if $T_{min} = T_1$ and $x^{l_n}_{min} := x^{l_n}_{o}$ otherwise. The time series $x^{l_n}_{min}$ remains unchanged. To make the other modality's SITS the same length, we adopt nearest-timestamp interpolation: for each image ${x^{l_n}_{min_i}} \in x^{l_n}_{min}$, we find the image in the corresponding time series of the opposite modality that was captured at the time closest to ${x^{l_n}_{min_i}}$. The result of our interpolation strategy results in $N$ SITS pairs $\{(x^{l_n}_{o}, x^{l_n}_{r})\}^N_{n=1}$ where both modalities' time series each contain $T_{min}$ images coarsely aligned in time.

\para{Encoder Design }The first part of \name\ consists of an encoder network that takes the spatially aligned optical and radar SITS, $x^{l_n}_o$ and $x^{l_n}_r$, as input. The encoder consists of four convolution layers, of which the first two are input-specific based on modality. Let $y_r, y_o$ denote the first two layers (used for the radar and optical domains, resp.) and $y_c$ denote the last two layers. The encoder for the radar and optical domains are $f_r = y_c \circ y_r$ and $f_o = y_c \circ y_o$. We use the outputs of $f_r, f_o$ as the features passed to the rest of the network. 
The encoders $f_r(\cdot)$ and $f_o(\cdot)$ use a 3D U-Net backbone architecture~\cite{unet3d} consisting of convolution, batchnorm \cite{batchnorm}, and max-pooling layers with leaky ReLU activations. The 3D operations are applied along both the temporal dimension and the spatial dimensions. This architecture has been used as a state-of-the-art benchmark for a wide variety of prior work in SITS segmentation tasks~\cite{dynamicearthnet,ghanadataset,panop_segmentation} and offers a relatively simple design that is comparable in performance to other state-of-the-art SITS segmentation architectures that use a separate sequential technique to handle the temporal dimension.

\begin{figure}[t]
    \centering
    \includegraphics[width=.6\linewidth]{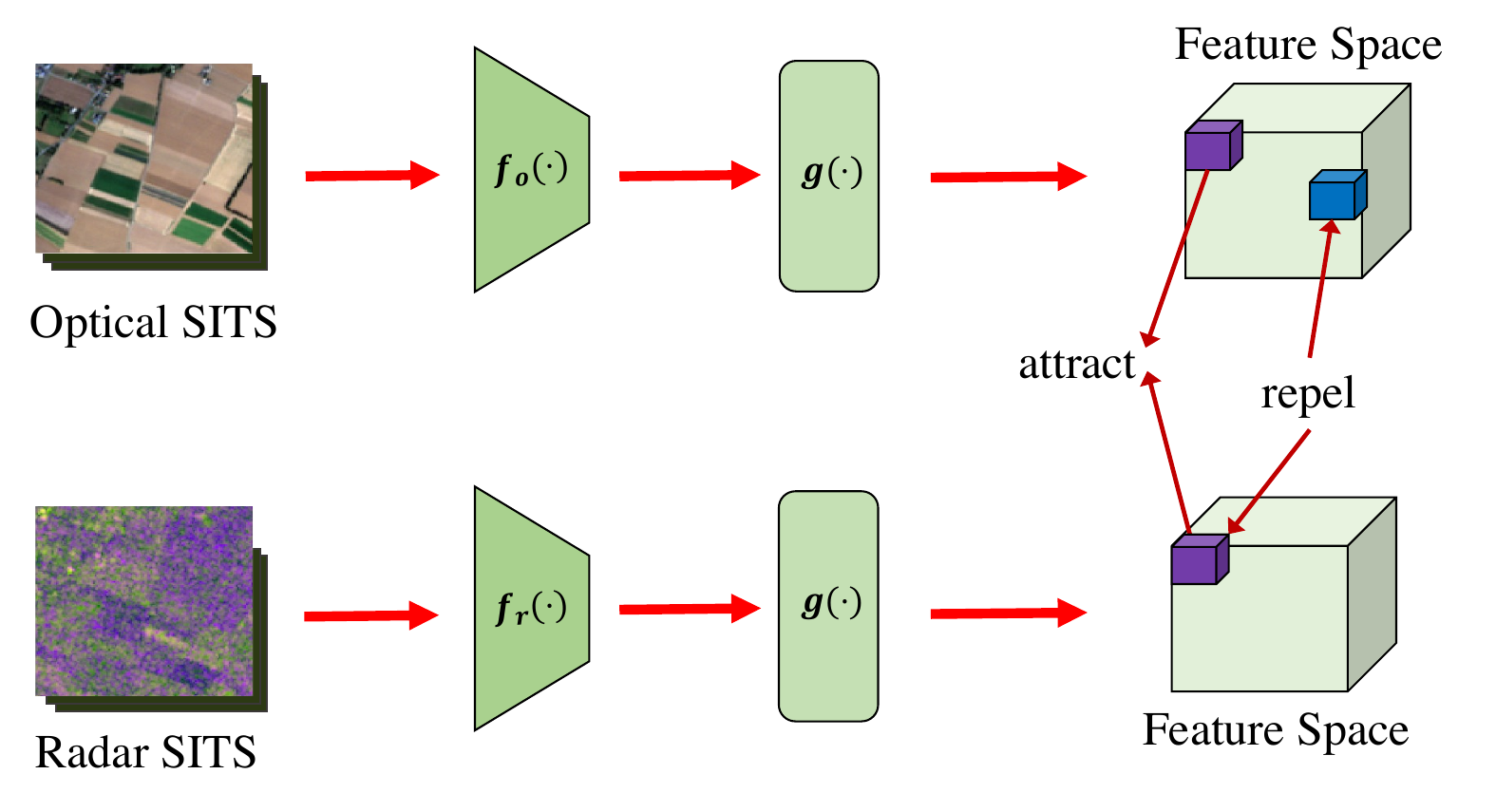}
    \caption{\textbf{Multi-Modal Space-Time Contrastive Learning for SITS.} Our approach operates on the encoded SITS feature maps. Corresponding space-time pixels on the feature map are denoted as positive pairs that the contrastive loss tries to align. Non-corresponding pixel pairs are negative and repelled by the loss. }
    \label{fig:contrastivelearning}
    \vspace{-0.2in}
\end{figure}
\subsection{Multi-Modal Space-Time Contrastive Learning}\label{contrastivesection} 

Our approach builds upon the key idea of \emph{semantic scene consistency between varying satellite modalities that are captured over the same space and at the same time}. Therefore, our encoder should map image pixels captured over similar space-time to similar representations; while encoding random non-corresponding pixels to differing representations. We incorporate this intuition in our training scheme through contrastive learning. Inspired by recent successes of pixel-wise contrastive learning \cite{pixcl}, we propose a pixel-wise contrastive loss which preserves the spatio-temporal structure of  our representations for better transfer to downstream pixel-level tasks like semantic segmentation. Fig.~\ref{fig:contrastivelearning} outlines this approach. 

Prior work~\cite{simclr,seco,gssl} on contrastive learning for images often use single image views and perform a variety of data augmentations (e.g., crop, rotate, blur) on a single view. Different augmentations that correspond to the same view are often correlated together as a positive pair for the loss function. However, in the case of satellite images, we benefit from the availability of multi-modal data and omit the augmentation step. Each modality captures the same view of Earth at different wavelengths, and can be used as a different transformation.

Similar to prior work \cite{simclr,pixcl}, we implement a projection head network $g(\cdot)$ that maps the output of $f_r(\cdot)$ and $f_o(\cdot)$ to the latent space where the contrastive loss is applied. The projection head consists of two successive  $1 \times 1 \times 1$ 3D convolution layers with batch normalization and LeakyReLU activation. Note that $g(\cdot)$ is only used during contrastive pre-training and not while training the reconstruction network or the downstream segmentation network. The output of $g(\cdot)$ is a feature map of the encoded SITS with compressed spatio-temporal dimensions. 

We assign positive pairs as pixels in the feature space with the \textit{same spatial and temporal dimensions}, across different modalities. Pixels pairs with different space or time dimensions are considered as negative pairs in our loss, since they correspond to different semantics. 
We opt to use the InfoNCE loss\cite{infonce} as our contrastive loss function:
\begin{equation}
    \mathcal{L}_{c} = - \log\left(\frac{e^{\mathit{sim}(z_i,z_j)/\tau}}{e^{\mathit{sim}(z_i,z_j)/\tau}+\sum\limits_{z_n \in Z}e^{\mathit{sim}(z_i,z_n)/\tau}}\right)
\end{equation}
Positive pairs $z_i$ and $z_j$ are corresponding space-time pixels in the feature map representations of opposite modalities. $Z$ is the set of all negative feature map pairs with anchor pixel $z_i$ in the opposite modality. More broadly, $Z$ consists of the features that were captured at different spatial locations or different times from $z_i$. The cosine similarity function is defined as  $\mathit{sim}(u, v) = u^T v/\|u\| \|v\|$. The temperature hyperparameter $\tau$ is set to 0.5 by default. The loss is first averaged over all pixels in the first modality's feature map; then, we compute the loss averaged across all pixels in the second modality feature map as anchor pixels. Finally, we average the loss across both modalities together to compute the final contrastive loss per sample in the batch. 

\subsection{Cross-Modality Reconstruction Network}
\label{reconstructionsection}
Although acquiring semantic segmentation labels for SITS is challenging, an advantage of SITS is that images can be easily aligned spatially. To leverage the spatial alignment between multiple modalities, we design a reconstruction network that infers the SITS of one modality given the other. By learning to reconstruct SITS from other modalities as an auxiliary task, the reconstruction network is able to learn representative features for the input modality that are helpful for the downstream segmentation task.  

Our reconstruction network uses encoder $f_{in}$ (either $f_r$ or $f_o$ depending on the inference modality) and decoder $h$. The network takes as input a SITS from one modality (denoted $x^{l_n}_{in}$, which has $C_{in}$ channels) and attempts to reconstruct the corresponding SITS of the other modality (denoted $x^{l_n}_{out}$, which has $C_{out}$ channels). The output of our reconstruction network is the estimated reconstruction of the SITS of the second modality: $\hat{x}^{l_n}_{out} = h(f_{in}(x^{l_n}_{in}))$. We define the loss for our reconstruction network as the mean $L_1$ loss between the original and the reconstructed time series:
\begin{equation}\label{recons_loss}
    \mathcal{L}_{r} = \frac{\lVert \hat{x}^{l_n}_{out} - x^{l_n}_{out} \rVert_1}{T_{min} \cdot C_{out} \cdot H \cdot W} 
\end{equation}

\subsection{Downstream Training} \label{downstreamtraining}
Finally, after pre-training our network with spatially aligned modalities, we fine-tune the network on a small number of labeled samples. We use the same encoders $f_{in}(\cdot)$ and decoder $h(\cdot)$ networks used during pre-training. However, we modify the number of channels of the decoder's final convolution layer for the relevant segmentation map output. We carry out the downstream training using standard cross-entropy loss. 

\para{Generalizing to Other Temporal Encoders} A key advantage of \name\ is that it can be easily extended to other types of SITS segmentation architectures that may encode the temporal dimension differently. Such architectures may use convolutional layers to encode the spatial dimensions and a temporal model, such as LSTM/RNN \cite{rnn1,panop_segmentation}, to handle the temporal dimensions. In these cases, \name\ can first be used to train the convolutional spatial encoders of the network. During downstream training, the temporal encoder can be added to the network and trained using multi-modal features extracted from the spatial encoders.

\vspace{-.1in}
\section{Experiments}
\vspace{-.1in}
In this section, we describe experiments conducted to evaluate \name. We train all self-supervised models in two phases. First, we pre-train all models for 100 epochs. For pretraining, the models are trained using \datasetname, our curated geographic specific, pre-train datasets. The pre-train datasets consist only of images and do not have annotated labels. In the second stage, we fine-tune the network for the downstream segmentation task for 50 epochs using the datasets with annotated labels. For optical imagery, we train using only using the RGB channels to be consistent and fair to prior work in self-supervised models for remote sensing \cite{gssl,seco,caco}. Additional training details are in the supplementary material. 

\subsection{Pre-Train Datasets}
We demonstrate the efficacy of multi-modal self-supervised pre-training by gathering a large dataset of aligned optical and radar images. Our main motivation for curating this data set is to illustrate how geographic specific aligned multi-modal satellite data is easy to acquire, allowing for greater opportunities to benefit from pre-training. Given the geo-tagged characteristic of satellite images, we can also collect data from geographic specific locations and study how the geographic location of images from certain regions can have an impact on the performance of downstream segmentation. Images in our dataset were collected from Sentinel 1 and 2 satellites and were aligned using the Microsoft Farmvibes Space Eye work flow \cite{MicrosoftFarmVibesAI}. We collect datasets for about a year for each corresponding region of our fine tune datasets. Specifically, we collect a pre-training data set that contains 5314 time series with a total of 731k Sentinel 1 images and 90k Sentinel 2 images for PASTIS-R dataset (described below). For Africa crop type mapping, our dataset contains 5941 time series, with 193k Sentinel 1 images and 70k Sentinel 2 images. We plan to make these datasets publicly available.

\subsection{Fine Tune Datasets}
\noindent\textbf{PASTIS-R:} The PASTIS-R \cite{pastis-r} agricultural dataset contains 2,433 optical and radar SITS from the ESA's Sentinel 1 and 2 satellites. Each SITS contains between 38 and 61 images taken between September 2018 and November 2019. The dataset provides an annotated semantic segmentation map for each of 2,433 spatial locations, where every pixel is given a semantic label from one of 20 different crop type classes. Many optical images are partially occluded by clouds. Note that we only consider the semantic segmentation labels from this data set and DO NOT perform parcel classification experiments (as done in the original paper) as semantic segmentation is a strictly more challenging task.

\noindent\textbf{Africa Crop Type Mapping:} The Africa Crop Type Mapping dataset \cite{africacrop} contains multi-modal SITS over various regions in Africa. Ground truth labels in this dataset were collected for 4 classes in 2017. For our experiments, we use  837 fields in the South Sudan partition.

\setlength\tabcolsep{10pt}
\begin{table}[t]
\centering
\caption{Segmentation results on the PASTIS-R and Africa Crop Type Datasets (mIoU).}
{%
\setlength\tabcolsep{1.5pt}
\begin{tabular}{p{.6in}||p{.4in}p{.4in}p{.4in}p{.4in}||p{.4in}p{.4in}p{.4in}p{.4in}}
\toprule
Dataset&\multicolumn{4}{c||}{PASTIS-R}&\multicolumn{4}{c}{Africa Crop Type}
\\\midrule
Method & Radar 100\% & Optical 100\% & Radar 10\% & Optical 10\% & Radar 100\% & Optical 100 \% & Radar 10\%   & Optical 10\% \\
\midrule
SatMAE & 37.9 & 36.2 & 11.1 & 28.2& 12.4 & 11.9 & 3.30 & 7.06 \\
SeCo &43.3 & 23.9 &27.2 & 12.4& 9.05 & 9.05 & 2.85 & 2.58\\
GSSL & 41.8 & 21.9 & 30.9 & 13.6 & 18.8 & 4.36 & 5.67 & 8.04\\
CaCO &42.7 & 23.4 & 24.9 & 16.4& 9.05 & 9.05 & 2.78 & 2.82 \\
MMF\footnote{MMF stands for Multi-Modal Fusion} & 53.2 & 48.3 & 36.0 & 27.4 & 9.08 & 18.1 & 9.02 & 18.5\\
\midrule
\textbf{\name\ (Ours)} & \textbf{54.6} & \textbf{53.4} & \textbf{36.5} & \textbf{33.7} &  \textbf{21.2} &  \textbf{31.2} &  \textbf{10.2} &  \textbf{24.4}\\
\bottomrule
\end{tabular}
}
\label{Table_PASTIS}
\end{table}

\subsection{Baselines}
We benchmark \name\ against several competing self-supervised baselines. To the best of our knowledge, we are the first self-supervised approach that leverages multi-modal imagery for SITS.

\textit{SatMAE:} SatMAE \cite{satmae} is a SOTA masked AutoEncoder based method designed specifically for multi-temporal satellite imagery. The method uses a multi-temporal vision transformer architecture. Note that since we are dealing with SITS, we compare against the multi-temporal version of SatMAE.

\textit{Contrastive Methods:} We compare against modern prior self-supervised work for remote sensing that uses single-modal contrastive loss. We compare against SeCo \cite{seco}, GSSL \cite{gssl}, and CaCo \cite{caco}, all of which use contrastive learning to align single-modal image pairs. In their original papers, these baselines use a single image ResNet-NN network to contrast scenes of different timestamps. We augment them with a ConvLSTM decoder to integrate over the temporal dimension for fair comparison.

\textit{Self-Supervised Multi-Modal Fusion:} This approach is a naive self-supervised approach of leveraging multi-modal data for SITS segmentation. Let $x^{l_n}_{m_1}$ denote the SITS modality we have access to at inference time. We first pre-train a network $r(\cdot)$ that, given $x^{l_n}_{m_1}$, learns to reconstruct the SITS of the other modality $x^{l_n}_{m_2}$ (using the loss in Eq.~\ref{recons_loss}). Then, we train a separate network that takes as input the concatenation of $x^{l_n}_{m_1}$ and $r(x^{l_n}_{m_1})$. Using this network, we produce the segmentation label using the PASTIS early fusion technique \cite{pastis-r}. During inference, we similarly generate the SITS of the missing modality using the reconstruction network and perform segmentation on the original and generated modalities.

\subsection{Quantitative Segmentation Results}
We first examine the segmentation performance quantitatively using only a few labels for downstream training.

\para{Results on PASTIS-R:} Table~\ref{Table_PASTIS} reports the mIoU on the PASTIS-R test set using both 100\% and 10\% of the labeled dataset. \name\ outperforms all competing baselines across the board for both optical and radar inference experiments. We observe less relative improvement in the radar inference experiments due to radar being a low-resolution modality that provides less information than non-occluded optical images. We also observe greater improvement when the amount of labeled data provided is lower. Finally, although the self-supervised fusion technique leverages multi-modal, temporal data and largely outperforms all other baselines, \name\ provides greater performance gain through its sophisticated cross-modal contrastive and reconstruction framework.

\noindent\textbf{Results on Africa Crop Type Mapping:} In Table~\ref{Table_PASTIS}, we report segmentation results on the Africa Crop Type Mapping test set. This dataset is more challenging due to multiple reasons. First, the pre-train dataset contains less temporal information due to sparse image collection by the Sentinel satellites. Therefore, we see lower mIoU values for this dataset. However, \name\ continues to significantly outperform all baselines for different modalities. Without \name's self-supervised multi-modal approach, the mIoU drops for both single and multi-model baselines.

\begin{figure*}[t]
\centering
    \begin{subfigure}{.45\textwidth} 
        \centering
        \caption{Ablation between our two proposed losses (mIoU). Inf. is short for Inference Modality.}
        \raisebox{.62in}{\begin{tabular}{ccc}
        \toprule
        Ablation & Inf. Type & mIoU\\
        \midrule
        $\mathcal{L}_{c}$ & Optical &  52.6\\
        $\mathcal{L}_{r}$ & Optical &  52.5\\
        $\mathcal{L}_{c}$ & Radar & 53.6\\
        $\mathcal{L}_{r}$ & Radar & 54.6\\
        \midrule
        $\mathcal{L}_{r} + \mathcal{L}_{c}$ & Optical & \textbf{53.4}\\
        $\mathcal{L}_{r} + \mathcal{L}_{c}$ & Radar & \textbf{54.6}\\
        \bottomrule
        \end{tabular}}
        \label{LossAblation}
    \end{subfigure}
    \hfill 
    \begin{subfigure}{.5\textwidth} 
        \centering
        \caption{Ablation on modalities. Inf. is short for Inference Modality}
        {\footnotesize
        \begin{tabular}{lcc}
        \toprule
        Ablation & Inf. Type & mIoU \\
        \midrule
        Single Modal & Optical & 51.3 \\
        Single Modal & Radar & 53.8 \\

        \midrule
        \name\ (Multi-Modal) & Optical &\textbf{53.4}\\
        \name\ (Multi-Modal) & Radar &\textbf{54.6}\\

        \bottomrule
        \end{tabular}}
        \label{Ablationmodality}
    \end{subfigure}
    \caption{Loss and modality ablation for \name.}
\end{figure*}

\begin{figure*}[t]
\centering
    \begin{subfigure}[b]{.35\textwidth} 
        \centering
        \caption{Ablation on geographic location. SL is short for Same Location pre-training. DL is short for Different Location pre-training.\label{Ablation}  \label{fig:geographicablation}}
        \setlength\tabcolsep{5.5pt}
        \raisebox{1.2 in}{\begin{tabular}{lcc}
        \toprule
        Ablation & Inf. Type & mIoU\\
        \midrule
        SL 10\% & Optical &  33.7\\
        SL 10\% & Radar &  36.5\\
        SL 100\% & Optical & 53.4\\
        SL 100\%& Radar & 54.6\\
        \midrule
        DL 10\% & Optical &  29.5\\
        DL 10\% & Radar &  33.4\\
        DL 100\% & Optical & 48.7\\
        DL 100\%& Radar & 51.9\\
        \bottomrule
        \end{tabular}}

    \end{subfigure}
    \hspace{.4in}
    \begin{subfigure}[b]{.4\textwidth}
    \centering
    {%
        \includegraphics[width=\linewidth]{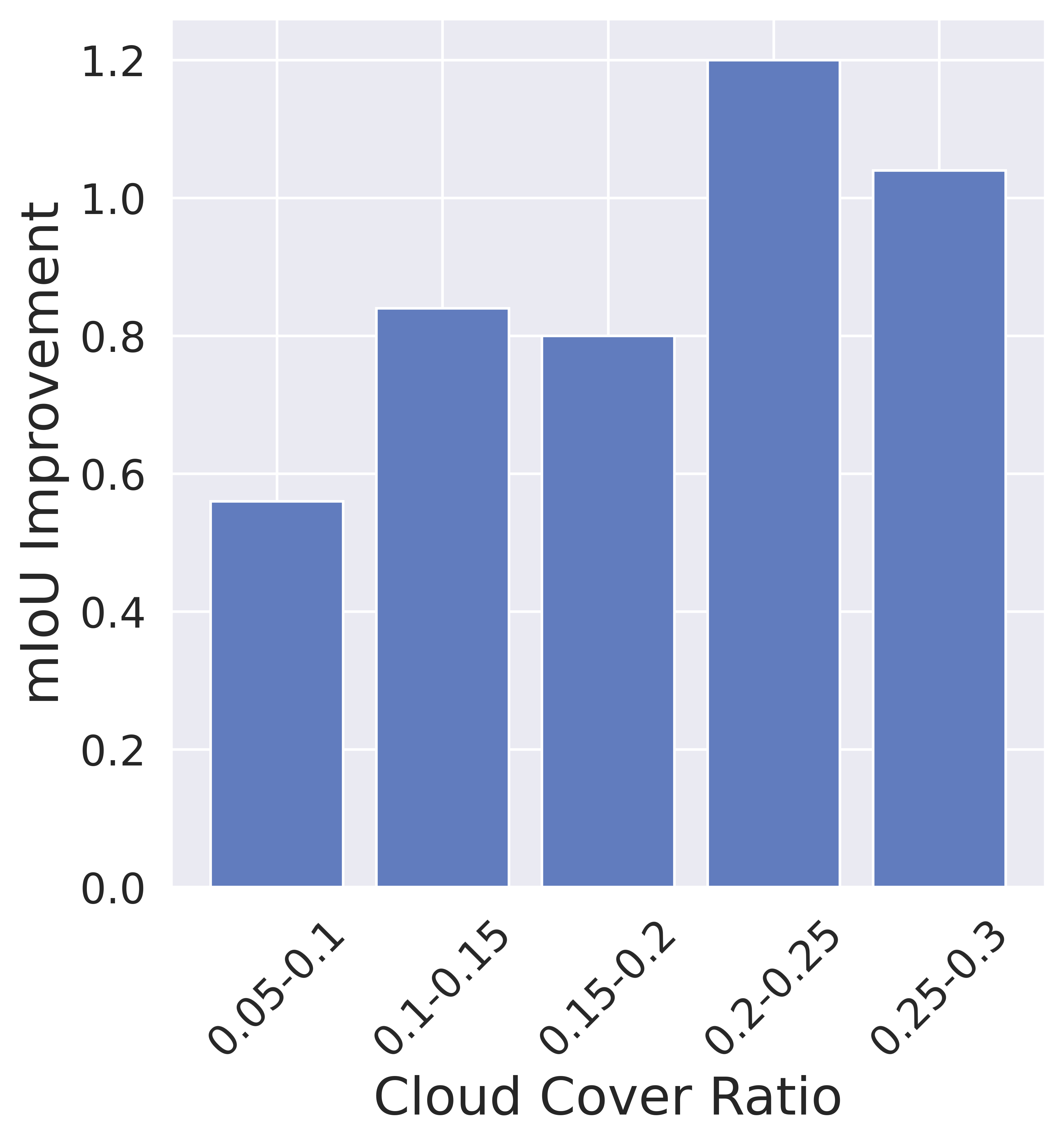}
        }
        \caption{Cloud cover robustness prediction.}
        \label{fig:cloudrobustness}
    \end{subfigure}
    \caption{Geographic ablation and cloud cover robustness results.}
\end{figure*}

\subsection{Ablation Study}
\vspace{-.09in}
We provide ablation studies using the PASTIS dataset. 

\noindent\textbf{Loss Ablation:} We measure the individual contribution of different losses used in \name. Table \ref{LossAblation} reports the results using the PASTIS dataset with both radar and optical inference. In both scenarios, the benefits of jointly optimizing the contrastive and reconstruction losses are higher as the number of labels increases. This demonstrates \name's ability to provide both temporal and spatial alignment benefits in pretraining to improve downstream model performance. 

\noindent\textbf{Modality Ablation:} We measure the effect of multiple modalities in Table~\ref{Ablationmodality}. We compare against a unimodal variant of \name, where our proposed contrastive objective operates over an optical SITS and the same optical SITS with random augmentations, similar to how contrastive loss is used in prior work. We find significant gains in performance when the radar modality is added during training. This holds true for both scenarios of inference modality.

\noindent\textbf{Geographical Ablation:} We report ablation results on the PASTIS dataset by pretraining on our curated, unlabeled Africa dataset. Table~\ref{fig:geographicablation} reports our results for SL (Same-Location Pretraining) and DL (Different-Location Pretraining). In this setup, we fix our pre-training sets to have the same number of SITS samples for fair comparison. Although we see a dip in performance due to different geographical pretraining locations, which is less representative of the finetune dataset, the performance drop is limited and still performs well compared to other self-supervised approaches even when they use SL pretraining. This demonstrates the utility of \name\ even on data where the geographic location is unknown.

\begin{figure*}[t]
    \centering
    \includegraphics[width=.9\linewidth]{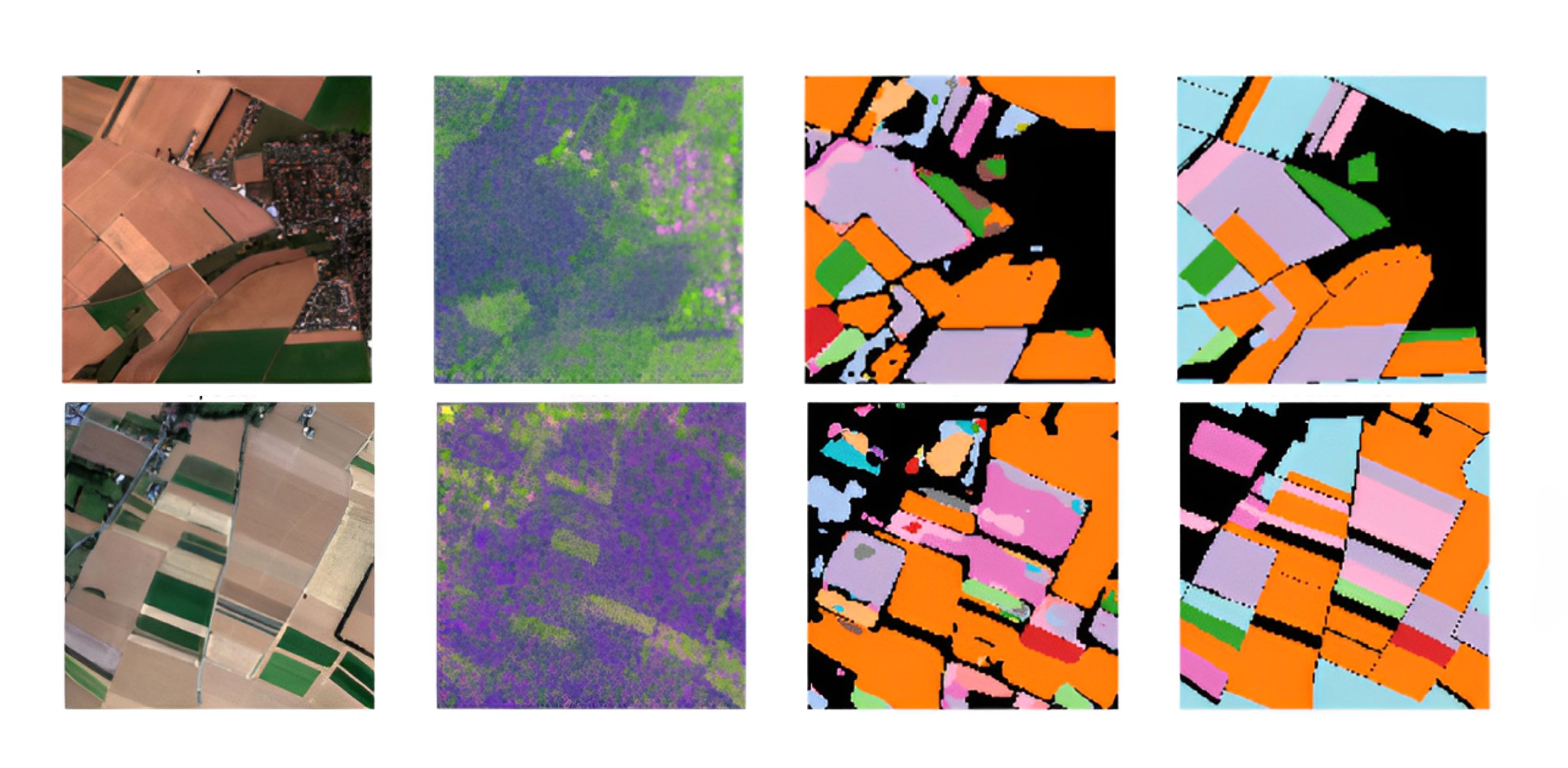}\vspace{-0.2in}
    \caption{Qualitative results on optical inference. Each row represents a different sample or geographic location from the PASTIS-R dataset for \name's evaluation. The first column (leftmost) is a single optical image from the optical SITS. The second column is a single radar image from the radar SITS. The third column is the prediction from \name. The fourth column (rightmost) is the ground truth segmentation map.  }\vspace{-0.2in}
    \label{fig:qualitativeresults}
\end{figure*}
\subsection{Qualitative Evaluation}
\vspace{-0.1in}
In Fig.~\ref{fig:qualitativeresults}, we plot an example of segmentation results of \name\ from the test set of PASTIS from models trained with 100\% of the labels. We find that our approach can segment even the hard class labels (one's with few training examples) e.g the patches in light green as shown in Fig.~\ref{fig:qualitativeresults}.

\subsection{Robustness to Cloud Cover}\vspace{-0.1in}
We analyze \name's ability to tackle the challenge of cloud cover during inference on optical images. 
We obtain a cloud mask for every image in the PASTIS test set using the S2Cloudless algorithm \cite{s2cloudless}. We compute the cloud cover ratio as the number of clouded pixels to total pixels in the SITS. After grouping every SITS by cloud cover ratio, we compute the mIoU. Fig.~\ref{fig:cloudrobustness} reports the mIoU gain of \name\ over the CaCo baseline for 100\% labels (the scenario when our model has the highest relative improvement). The mIoU improvement is the mean improvement over CaCo baseline. The results illustrate that our approach provides greater gains in the presence of clouds. The radar data can help guide the model to make better predictions on partially clouded data, since the radar images provide insights into how the model can ``see through the clouds. The improvement drops for SITS where the cloud cover ratio is greater than 25\%, where many images are mostly or even fully occluded by clouds (and contain little useful information).

\vspace{-.1in}
\section{Conclusion}
\vspace{-0.1in}
We introduced \name, a multi-modal self-supervised training framework for satellite image time series. \name\ uses a joint pixel-wise contrastive and reconstruction loss for multi-modal satellite imagery to improve segmentation performance on little labeled data. We envision \name\ will unlock the potential of using satellite imagery for emerging Earth-scale applications like climate monitoring and precision agriculture by reducing the requirement for large, labeled datasets.

%
%
\bibliographystyle{splncs04}
\bibliography{egbib}
\end{document}